\documentclass[default,iicol]{sn-jnl}

\usepackage{etoolbox}
\makeatletter
\patchcmd{\ps@headings}
{\def\@oddfoot{\hfill}%
	\let\@evenfoot\@oddfoot%
	\def\@evenhead{%
		\vbox to 0pt{\vspace*{-39pt}%
			\hbox to \hsize{\hfill Springer Nature 2021 \LaTeX\ template\hfill}}\par
		\hspace*{-\textwidth}\hbox to \hsize{\headerfont\thepage\qquad\rightmark\hfill}}%
	\def\@oddhead{%
		\vbox to 0pt{\vspace*{-39pt}%
			\hbox to \hsize{\hfill Springer Nature 2021 \LaTeX\ template\hfill}}\par
		\hspace*{-\textwidth}\hbox to \hsize{\headerfont\hfill\leftmark\qquad\thepage}}%
	\let\@mkboth\markboth}
{\def\@oddfoot{\hfill}%
	\let\@evenfoot\@oddfoot%
	\def\@evenhead{%
		\vbox to 0pt{\vspace*{-39pt}%
			\hbox to \hsize{}}\par
		\hspace*{-\textwidth}\hbox to \hsize{}}%
	\def\@oddhead{%
		\vbox to 0pt{\vspace*{-39pt}%
			\hbox to \hsize{}}\par
		\hspace*{-\textwidth}\hbox to \hsize{}}%
	\let\@mkboth\markboth}
{}
{}
\patchcmd{\ps@titlepage}
{\hbox to \hsize{\hfill Springer Nature 2021 \LaTeX\ template\hfill}}
{\hbox to \hsize{}}
{}
{}
\makeatother

\usepackage{subcaption}
\usepackage{siunitx}
\usepackage[shortlabels]{enumitem}
\usepackage{multirow}
\usepackage{amsmath}
\usepackage{graphicx}

\jyear{2022}%

\theoremstyle{thmstyleone}%
\theoremstyle{thmstyletwo}%

\theoremstyle{thmstylethree}%

\begin{document}

\title[Article Title]{Cascade Watchdog: A Multi-tiered Adversarial Guard for Outlier Detection}


\author*[1]{\fnm{Glauco} \sur{Amigo}}\email{Glauco\_Amigo1@Baylor.edu}

\author*[1]{\fnm{Justin M.} \sur{Bui}}\email{Justin\_Bui@Baylor.edu}

\author[1]{\fnm{Charles} \sur{Baylis}}

\author[1]{\fnm{Robert J.} \sur{Marks}}

\affil[1]{
	\orgname{Baylor University},
	\orgaddress{\city{Waco},  \state{TX}, \country{United States}}%
}


\abstract{The identification of out-of-distribution content is critical to the successful implementation of neural networks. Watchdog techniques have been developed to support the detection of these inputs, but the performance can be limited by the amount of available data. Generative adversarial networks have displayed numerous capabilities, including the ability to generate facsimiles with excellent accuracy. This paper presents and empirically evaluates a multi-tiered watchdog, which is developed using GAN generated data, for improved out-of-distribution detection. The cascade watchdog uses adversarial training to increase the amount of available data similar to the out-of-distribution elements that are more difficult to detect. Then, a specialized second guard is added sequentially. The results show a solid and significant improvement on the detection of the most challenging out-of-distribution inputs. 
}




\maketitle

\section{Introduction}

Data augmentation is described as  imagination or dreaming by C. Shorten and T.~M. Khoshgoftaar \cite{Shorten2019ASO}. They survey different methods of image data augmentation for deep learning, including adversarial training and generative adversarial networks (GANs) \cite{Goodfellow2014GAN}. Adversarial training can be used to attack or defend systems, as well as to increase the amount of available training data. The goal of data augmentation is to create new data samples from the existing training set. This new data is obtained according to the purposes of the application \cite{Marks1995boundary}. In this work, data augmentation serves to supplement the kind of out-of-distribution outlier data  that resides closest to the distribution manifold. 

In previous work \cite{Bui_Symbiotic_2021,Bui_Watchdog_2021}, the autoencoder watchdog is introduced to identify outliers in classification neural networks. The autoencoder watchdog measures an error function, such as the root mean square error (RSME), between the network's input and output. 
This is 
a measure of the distance between the autoencoder input and the training data manifold manifest in the latent space.

Data samples 
distant from the manifold are out-of-distribution. 
The closer the data gets to the manifold, the more fuzzy the classification becomes.
The work presented in this paper specializes in identifying outliers that are close to the manifold of the distribution. 
This is achieved in two steps. First, generating an augmented training data set that lies close to the boundary of the distribution manifold. Second, creating a binary classifier neural network that  specializes on  differentiating between in-distribution data and on-the-boundary data. 

The resulting cascade watchdog is comprised of two layers of defense against outliers. 
The first layer of defense of the cascade watchdog is the autoencoder layer, and the second layer of defense is a binary classifier layer.

\section{Background}
GANs are effective tools for data augmentation. Since the first publication, where Goodfellow et al. introduced GANs in 2014 \cite{Goodfellow2014GAN}, a variety of techniques and applications have been developed  across diverse fields.  Yi et al. \cite{Yi_2019}  present a review of adversarial training in medical imaging, one of the most prolific fields of application of GANs on data augmentation.

The basic structure of a GAN consists of a generator and a discriminator. The generator is trained to fool the discriminator, while the discriminator is trained to differentiate between real and fake inputs. After a GAN has been trained, it provides a source of freshly generated data samples which resemble real system inputs. Once the generator of the GAN is trained, it produces in-distribution samples from noise. 
Variations of GANS and their many applications are available in the literature \cite{Frid_Adar_2018,zhang2020gan,Zhu2018EmotionCW}.

This paper presents a method of adversarial training inspired by the GAN model. The main contrast between the generative adversarial training method presented here and other widespread applications of GANs resides in the different target. Normally, the goal of a GAN is to generate data in the distribution of the dataset. However, in this paper, the goal is to obtain out-of-distribution samples within a certain distance of the distribution. 

This is work based on autoencoder watchdog neural networks for outlier identification in classification networks \cite{Bui_Watchdog_2021,Bui_Symbiotic_2021}. The cascade watchdog improves upon the precision of the outlier identification task. It is capable of both identifying more outliers and reducing misidentification significantly. 
In a nutshell:

\begin{enumerate}
	\item The autoencoder serves as the discriminator of the GAN module,
	\item The GAN module generates out-of-distribution data samples close to the distribution manifold, and 
	\item A combination of in-distribution and generated out-of-distribution data is used to train the binary classifier.
\end{enumerate}

The binary classifier specializes in identifying outliers that are closer to the distribution manifold. Data far within the manifold are easy to classify. Classification of data close to the manifold surface is more difficult. Applying the autoencoder and the binary classifier in sequential order improves outlier identification, while preventing the network from discarding in-distribution elements by mistake.

Other approaches are available for outlier identification. Atlas et al. \cite{atlas1990X} and Hwang et al. \cite{hwang1990,hwang1991} identified manifold boundary points using neural network inversion \cite{jensen1997location,jensen1999inversion,thompson2003inversion}.
Yu et al. \cite{yu2019} propose a method that identifies outliers that are only far from the distribution. Lee et al. \cite{lee2018training} use a generator component to produce data samples on the boundary of the distribution. They train the classifier to assign less confidence to the classification of inputs on the boundary of the distribution. In order to obtain less confidence at the output of the classifier for 'boundary' inputs, they set the output target as the uniform distribution  for 'boundary' inputs during the training process.

\section{Methodology}
The distribution manifold 
lies is a small portion of the input space. The autoencoder layer is capable of identifying many outliers, covering a significant portion of the input space. To determine the boundary between in-distribution and out-of-distribution, the autoencoder uses a threshold hyperparameter. When selecting the threshold, there is  a trade-off between false negatives (out-of-distribution data samples that are not identified as outliers) and false positives (in-distribution data samples that are classified as outliers). A large threshold reduces the false positive rate but also increases the false negatives rate. Whereas a small threshold reduces the false negative rate but increases the false positive rate.

The second layer of the cascade watchdog is a fine grained binary classifier that  complements the autoencoder layer. The binary classifier specializes in identifying outliers that reside close to the distribution. The threshold of the autoencoder can be increased to reduce the false positive rate relying on the additional layer of defense provided by the binary classifier, which decreases the false negative rate safely. At the end of this process, both the false positive and false negative rates are reduced. More outliers are identified and less in-distribution data are erroneously discarded.

\subsection{Adversarial Watchdog}
The autoencoder also supports the development of a GAN, which generates out-of-distribution data samples close to the autoencoder threshold. For training  the GAN, two 
goals are necessary to generate a rich dataset: 

\begin{enumerate}
	\item Producing data samples where the autoencoder produces an error similar to the threshold.
	\item Distributing the generated data samples across the boundary of the manifold. To avoid the collapse of the GAN, each generated output depends on a point on the manifold taken from the in-distribution training set. Each generated data sample comes from one input on the training data. The distance between generated outliers and inputs must be similar to the threshold of the autoencoder. The same error function between the input and output of the autoencoder is used to measure the distance between the generated out-of-distribution data sample and the original in-distribution data sample.
\end{enumerate}

The combination of these two targets generates a dataset that is spread through the space near the boundary of the distribution manifold, preventing the collapse of the GAN.

The sequence of steps to produce the second layer of the cascade watchdog is:

\begin{enumerate}
	\item Train the autoenoder.
	\item Train the GAN and generate the dataset on the boundary of the distribution.
	\item Create a fine grained binary classifier.
\end{enumerate}

The last step consist of training the binary classifier in two categories: in-distribution and out-of-distribution. The original training dataset is labeled as in-distribution while the dataset on the boundary of the distribution generated with the  GAN is labeled as out-of-distribution.

\begin{figure}[!t]
	\centering
	\includegraphics[width=0.6\linewidth]{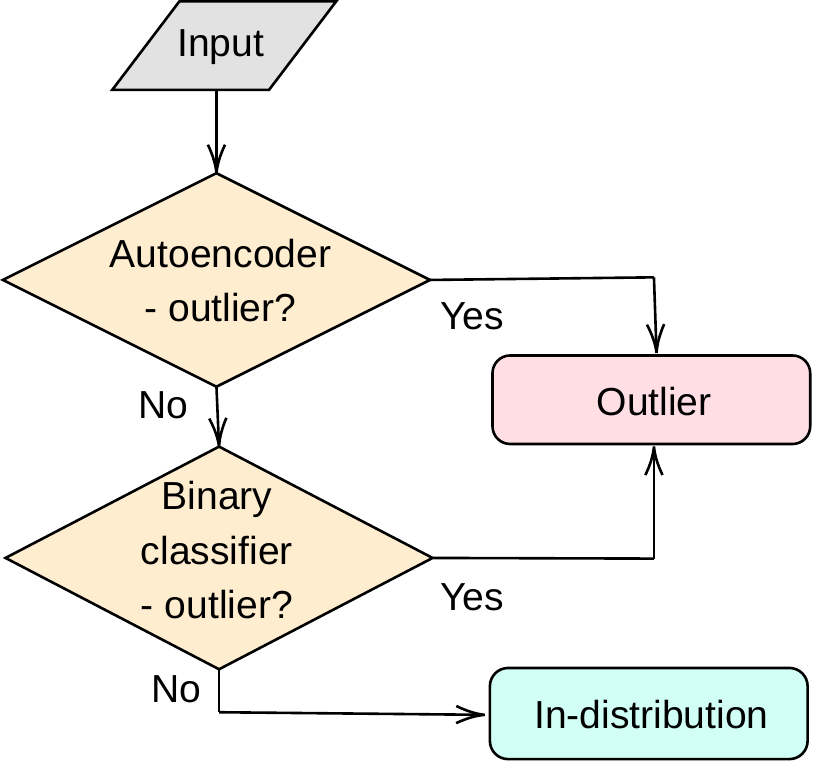}
	\caption{Flowchart of the cascaded watchdog. An input that passes both layers of defense is considered as pertaining into the distribution.}
	\label{fig:flowchartwd}
\end{figure}

After both the autoencoder and binary classifier are trained, they combine sequentially to form the cascade watchdog, as seen in Figure \ref{fig:flowchartwd}. First, the input is analyzed using the autoencoder layer, which identifies whether the input is an outlier or not. If the autoencoder does not identify the input as an outlier,  the input is then analyzed by the binary classifier. If  neither the autoencoder nor the binary classifier identifies the input as an outlier, then the input is considered to be an in-distribution element.

\begin{figure}[!t]
	\centering
	\includegraphics[width=1\linewidth]{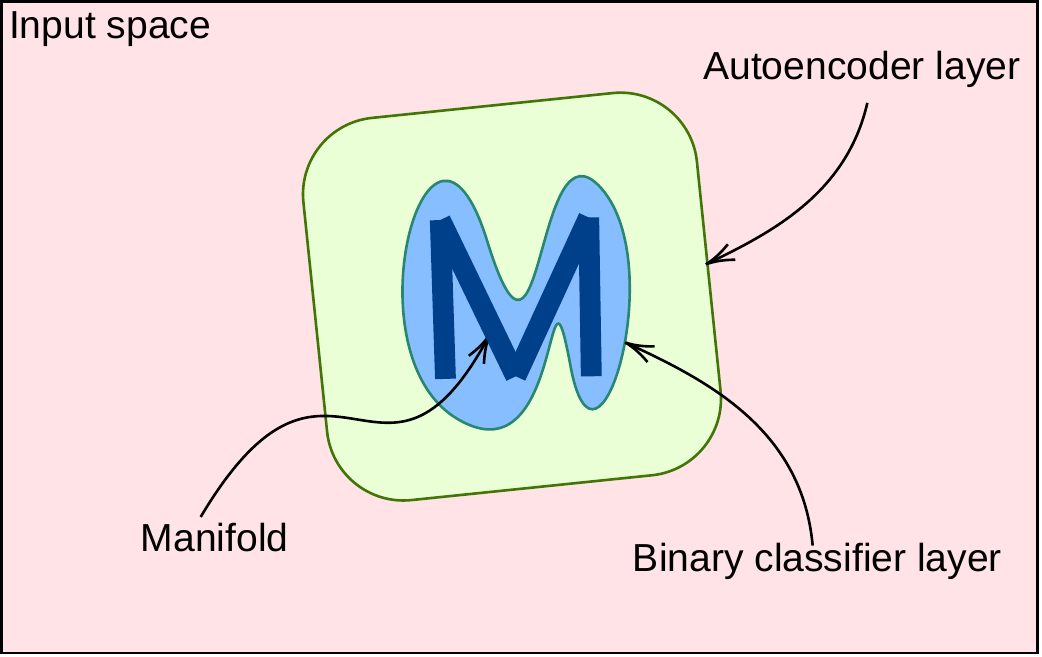}
	\caption{Venn diagram dividing the input space. The biggest area is the set of outliers detected by the autoencoder, inside  are those filtered by the binary classifier, and the bold ``M'' is the manifold with some false negatives around.}
	\label{fig:manifold}
\end{figure}

The performance of the binary classifier is evaluated with a ten-fold bias-variance analysis.  The quality of the binary classifier is measured by observing the  false and true negatives. Figure \ref{fig:manifold} shows a Venn diagram of the cascade watchdog formed by the autoencoder and the binary classifier. The domain of the autoencoder is the full input space and  the outliers that are farther from the manifold are detected, while the domain of the binary classifier is only the space that the autoencoder does not filter. Ideally, all of the outliers are detected while all in-distribution data is permitted. Observe that Figure \ref{fig:manifold} does not contain false positive outliers (i.e. no part of the manifold is marked as out-of-distribution), but  has false negative outliers (outliers that are very close to the manifold are not detected).

\section{Experimentation}
\begin{figure*}[!t]
	\centering
		\begin{subfigure}[t]{0.7\textwidth}
			\includegraphics[width=\textwidth]{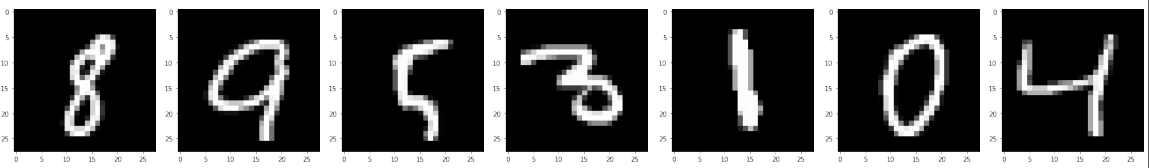}
			\caption{Original data samples from the MNIST dataset.}
			\label{fig:indistribution}
		\end{subfigure}

		\hfill

		\begin{subfigure}[t]{0.7\textwidth}
			\includegraphics[width=\textwidth]{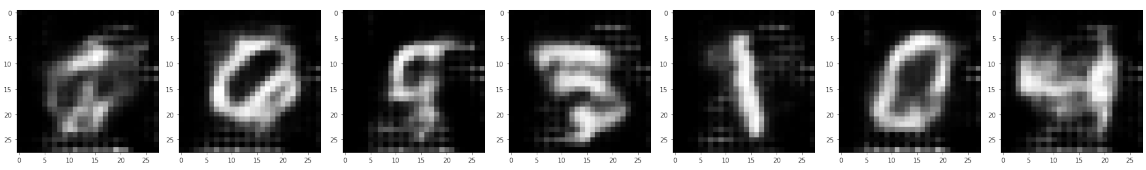}
			\caption{GAN generated on the boundary samples.}
			\label{fig:ontheboundary}
		\end{subfigure}
	
	\caption{Illustration of the boundary of the distribution. Each column corresponds to the input and the corresponding output of the GAN.}
	\label{fig:GANsamples}
\end{figure*}

The first step of the experiments is to train the autoencoder with the MNIST dataset%
\footnote{The MNIST dataset is available on TensorFlow: \url{https://www.tensorflow.org/datasets/catalog/mnist}}.
Examples of the MNIST dataset are shown in Figure \ref{fig:indistribution}. The GAN is trained to generate data samples close to  the boundary of the distribution (see some examples on Figure \ref{fig:ontheboundary}).

The boundary of the distribution is approximated by the threshold error function between the input and the output of the autoencoder. The error function chosen for the threshold of the autoencoder is the RMSE. For the experiments a value of 5 is selected for the autoencoder threshold. 

The loss function of the GAN has four components:

\begin{equation}\label{eq:lossfn}
	\text{GAN loss function} = \frac{\text{(a)}+\text{(b)}+\text{(c)}+\text{(d)}}{4}
\end{equation}
where:
\begin{enumerate}[(a)]
	\item $= \mid $threshold - RMSE(input, GAN output)$\mid$
	\item $=$ max\{0, RMSE(input, GAN output) - threshold\}
	\item $=\mid$threshold - RMSE(GAN output, autoencoder(GAN output))$\mid$
	\item $=$ max\{0, RMSE(GAN output, autoencoder(GAN output)) - threshold\}
\end{enumerate}
and:
\begin{itemize}[$\bullet$]
	\item ``$\mid$·$\mid$'' is the absolute value.
	\item Each ``input'' is a data sample from the MNIST training dataset.
	\item The ``GAN output'' is the data sample on the boundary of the distribution generated by the GAN from the current ``input'' that is being used for training.
	\item ``autoencoder(GAN output)'' is the reconstruction that the previously trained autoencoder produces from the ``GAN output''.
\end{itemize}
Note that the threshold hyperparameter in the loss function of the GAN is a different variable from the autoencoder threshold, even when they are technically related.

The loss function in \eqref{eq:lossfn} penalizes manifold distances that are greater or smaller than the threshold ---the GAN hyperparameter threshold. The distances to the manifold that are greater than the threshold (included on (a),(b),(c), and (d)) are penalized double than those that are less than the threshold (only included on (a) and (c)). 

The experiments reveal that  to generate images with a specific RMSE value on the autoencoder, the training of the GAN requires a smaller value for the threshold hyperparameter. For instance, the GAN needs a threshold hyperparameter of approximately 0.2 to obtain results with a RMSE of approximately 5.25;  a GAN threshold of 0.05 produces images with autoencoder RMSE values of about 1.3. Setting the hyperparameter of the GAN threshold about 0.3 or larger makes the GAN  collapse.

Using grid search \cite{lavalle2004relationship}, the optimal value for the threshold parameter of the GAN is $0.1375$. With this threshold, the GAN produces images with an average autoencoder RMSE of $4.14$. Even when the average RMSE of $4.14$ is below the threshold of the autoencoder value, set as~5, GAN generated images using this threshold as the distance to the manifold produce better results at the end of the experimentation pipeline.

After training the GAN, one data sample on-the-boundary of the distribution is obtained for each input of the original dataset (see  Figure \ref{fig:GANsamples}). The training dataset of the fine grained binary classifier layer consist of both the original in-distribution data and the generated on-the-boundary data. 

The training of the autoencoder is excluded from the bias-variance analysis. Once trained, the autoencoder is used for the 10 experimental runs. On each run, the GAN is trained reinitialized, allowing it to generate a fresh on-the-boundary dataset, then the binary classifier is zeroed and then retrained. \num{50000} data samples from the MNIST dataset are used in this process. 

In order to characterize the performance of the cascade watchdog,  \num{10000} samples from the  MNIST dataset and \num{10000} outlier samples  are used. The outlier samples are taken from the  Fashion MNIST dataset%
\footnote{The Fashion MNIST dataset is available on TensorFlow: \url{https://www.tensorflow.org/datasets/catalog/fashion_mnist}.}.
Samples of the Fashion MNIST dataset can be seen in Figure \ref{fig:examples_pictures}. The in-distribution samples from the MNIST dataset used for testing are separate from the samples used for training.

\begin{figure}[!t]
	\centering
	\begin{subfigure}[t]{0.5\textwidth}
		\centering
		\includegraphics[width=0.6\linewidth]{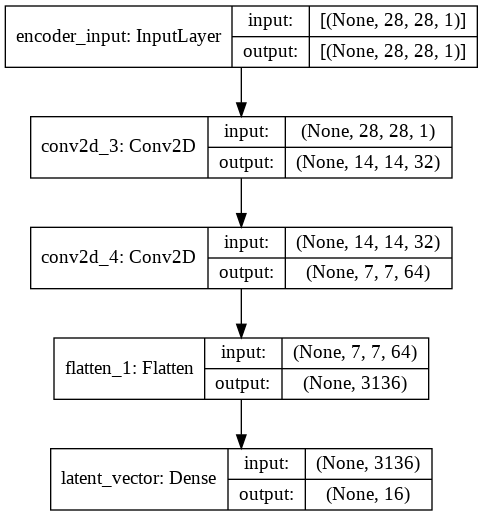}
		\caption{Encoder architecture.}
		\label{fig:encoderstructure}
	\end{subfigure}
	
	\hfill
	
	\begin{subfigure}[t]{0.5\textwidth}
		\centering
		\includegraphics[width=0.6\linewidth]{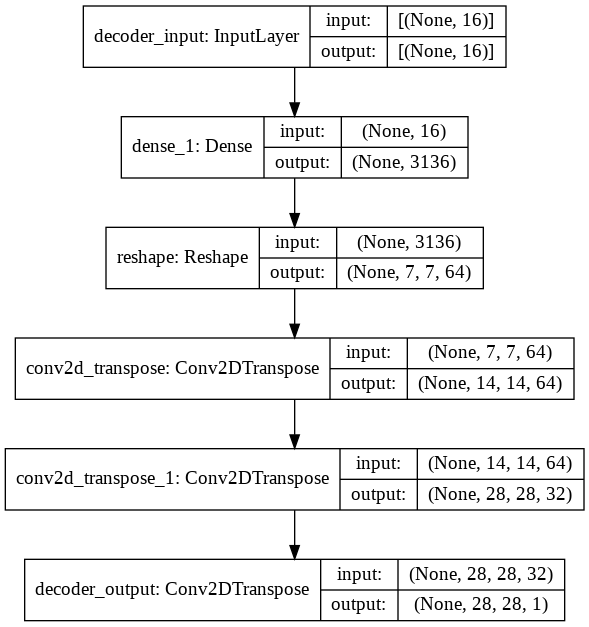}
		\caption{Decoder architecture.}
		\label{fig:decoder}
	\end{subfigure}
	
	\caption{Architecture of the autoencoder. The same structure is used for the autoencoder and for the GAN. Also the GAN uses transfer learning from the autoencoder for the encoder  and only trains the decoder.}
	\label{fig:AEarchitecture}
\end{figure}

The architecture of the autoencoder (see Figure \ref{fig:AEarchitecture}) has an encoder and a decoder connected sequentially. In between, the latent space has 16 features. The architecture of the GAN is the same as the autoencoder. Transfer learning is applied for the encoder block of the GAN by copying the  weighs from the autoencoder. The encoder of the GAN is then frozen during the training and only the decoder block of the GAN is trained. In essence, the autoencoder latent representation of the MNIST training dataset is used as input for the GAN decoder. Or, in other words,  the generative component of the GAN is trained to produce on-the-boundary data samples from the autoencoder latent representation of the in-distribution data samples.

\begin{figure}[!t]
	\centering
	\includegraphics[width=.6\linewidth]{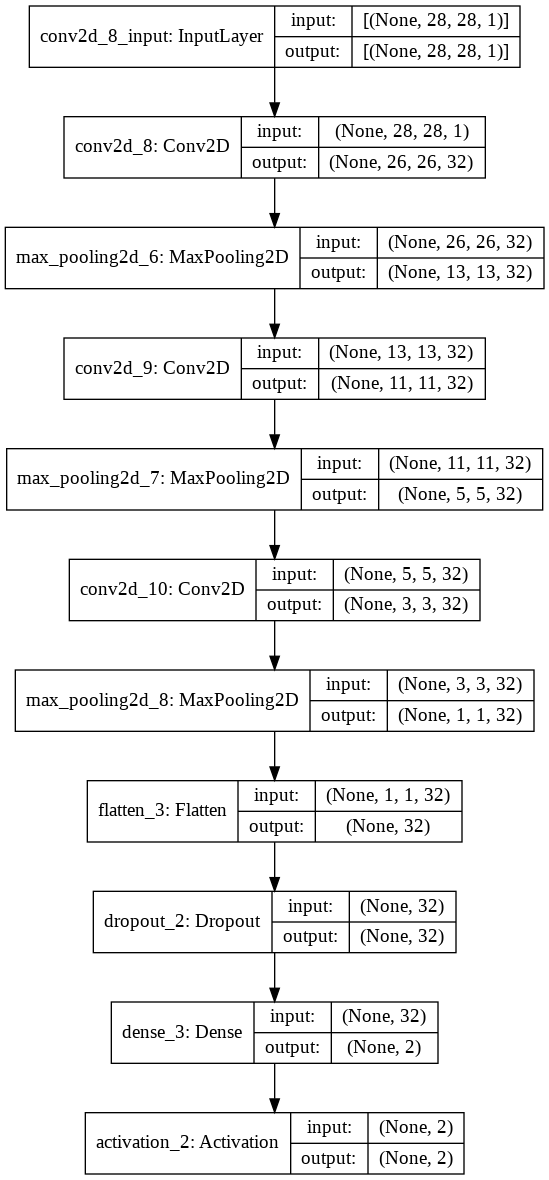}
	\caption{Architecture of the binary classifier used as the second layer defense of the cascade watchdog.}
	\label{fig:cnnbinarystructure}
\end{figure}

Finally, the architecture of the binary classifier is shown in Figure \ref{fig:cnnbinarystructure}. The binary classifier is a standard convolutional neural network classifier.

\section{Analysis}
The workflow of the cascade watchdog (see Figure \ref{fig:flowchartwd}) has two steps: 

\begin{enumerate}
	\item Autoencoder outlier detection.
	\item Binary classifier outlier detection.
\end{enumerate}

\begin{figure}[!t]
	\centering
	\includegraphics[width=1\linewidth]{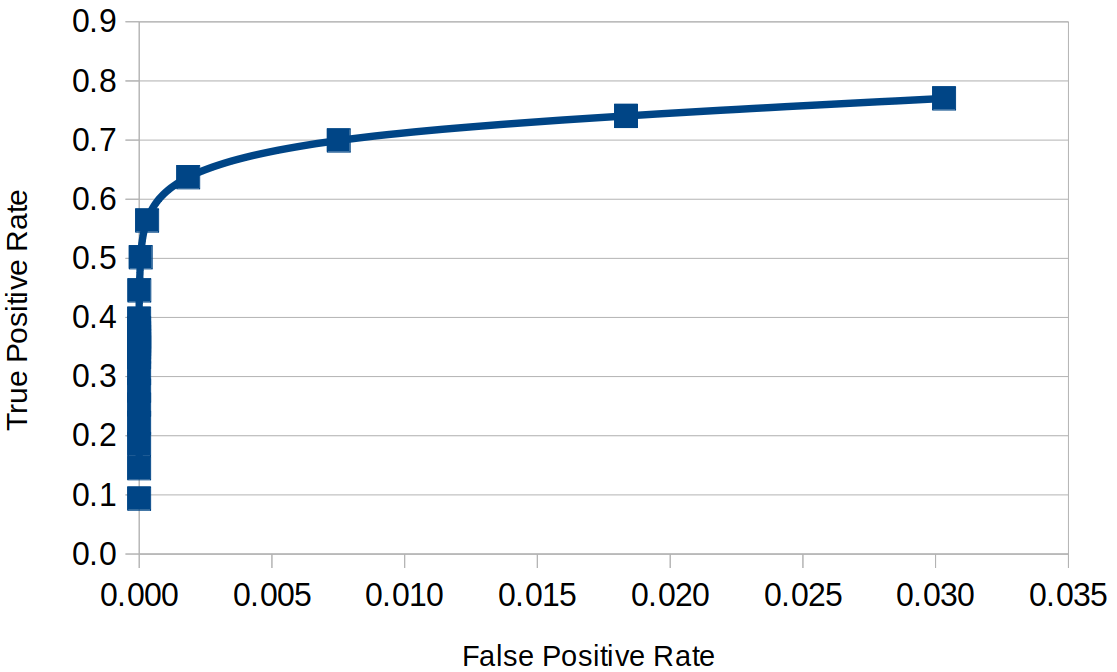}
	\caption{ROC curve for outlier detection by the binary classifier (see Table \ref{tab:roc}). The point $(1,1)$ and the diagonal, that would represent random guess, are  not included because they do not fit into the plot. }
	\label{fig:roc}
\end{figure}

In the first step, the autoencoder detected \num{9347} outliers out of \num{10000} data samples taken from the fashion MNIST dataset. In the second step, the remaining 653 outliers not detected by the autoencoder are tested on the binary classifier. The results of the ten fold bias-variance analysis are shown in Table \ref{tab:roc} and in Figure \ref{fig:roc}. In Figure~\ref{fig:roc}, the true positive rate accounts for the fraction of the outliers that the binary classifier detected, while the false positive rate corresponds to the in-distribution samples that the binary classifier layer classifies as outliers. The ROC curve characterizes a very good performance of the binary classifier, where the true positive rate increases to about $\frac{3}{4}$ while the false positive rate stays low. The specific values used to create this ROC curve are shown in Table \ref{tab:roc}.  The binary classifier has a true positive rate of 39.8\% with 9.5\% standard deviation while preserving a zero false positive rate. The false positive rate is 0 until the threshold of the binary classifier certainty is raised. Approaching the certainty threshold to the limit (certainty greater than 0\%) produces a small value for the false positive rate (3\%) while the true positive rate improves from 40\% to 77\% in average. 

\begin{table}[!t]
	\renewcommand{\arraystretch}{1.3}
	\caption{Ten-fold bias-variance analysis for outlier detection of the binary classifier. Note: The first column corresponds to the threshold applied to the output of  the softmax function on the binary classifier. }
	\label{tab:roc}
	\centering
	\includegraphics[width=0.47\textwidth]{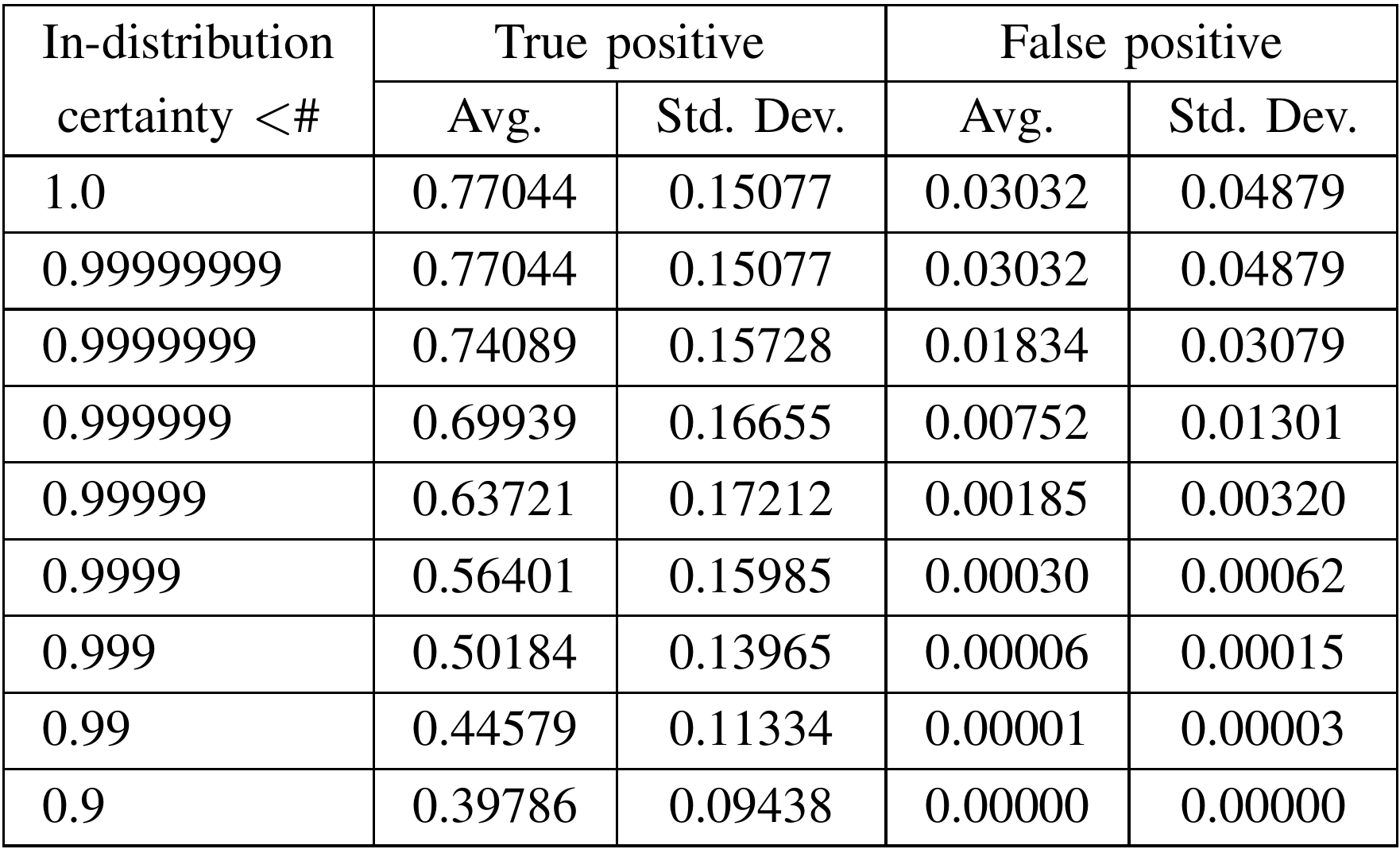}
\end{table}

\begin{figure}[!t]
	\centering
	
	\begin{subfigure}[t]{0.3\textwidth}
		\includegraphics[width=\textwidth]{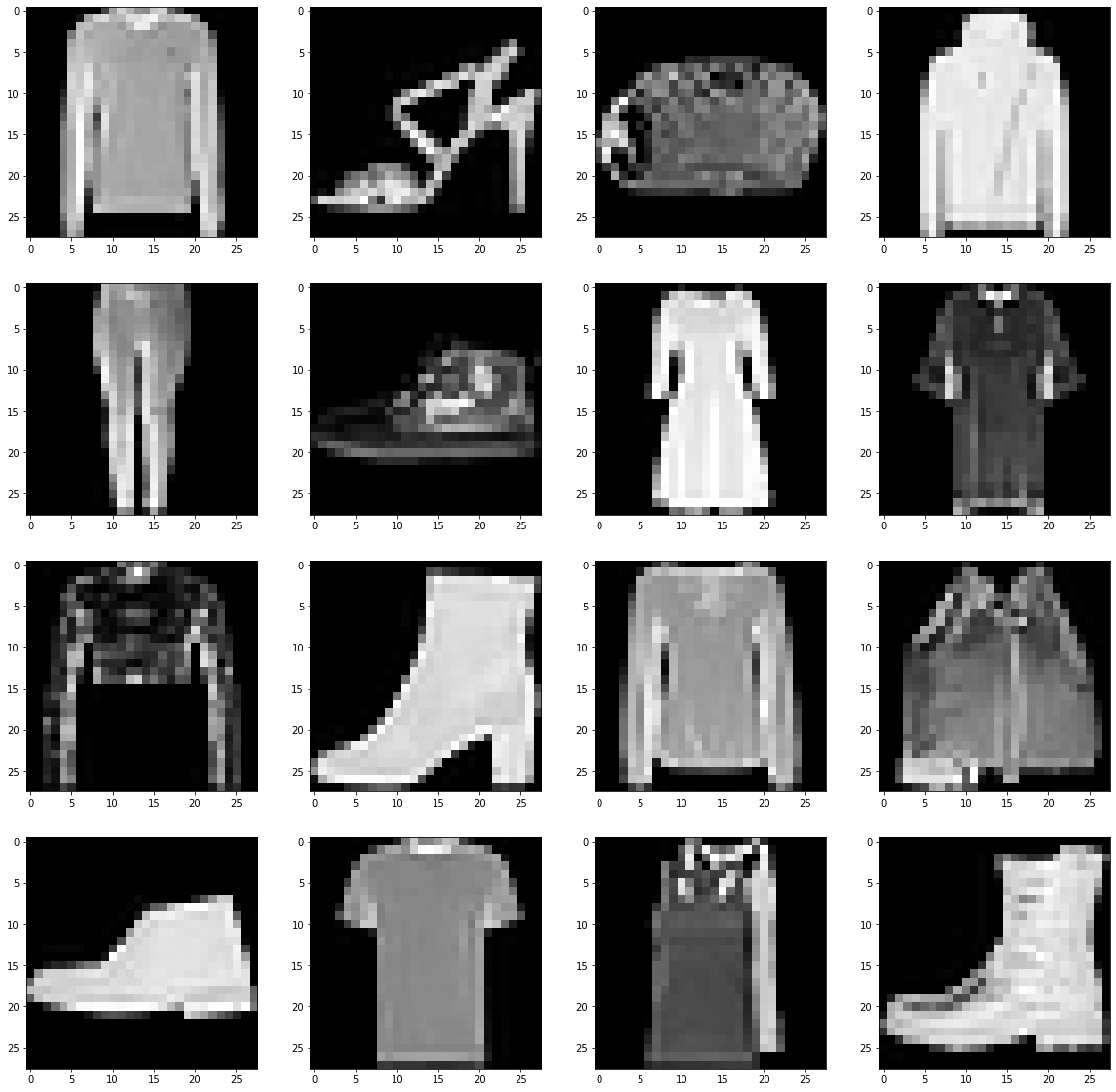}
		\caption{Outliers detected by the autoencoder layer.}
		\label{fig:excludedAELbig}
	\end{subfigure}
	
	\vfill
	
	\begin{subfigure}[t]{0.3\textwidth}
		\includegraphics[width=\textwidth]{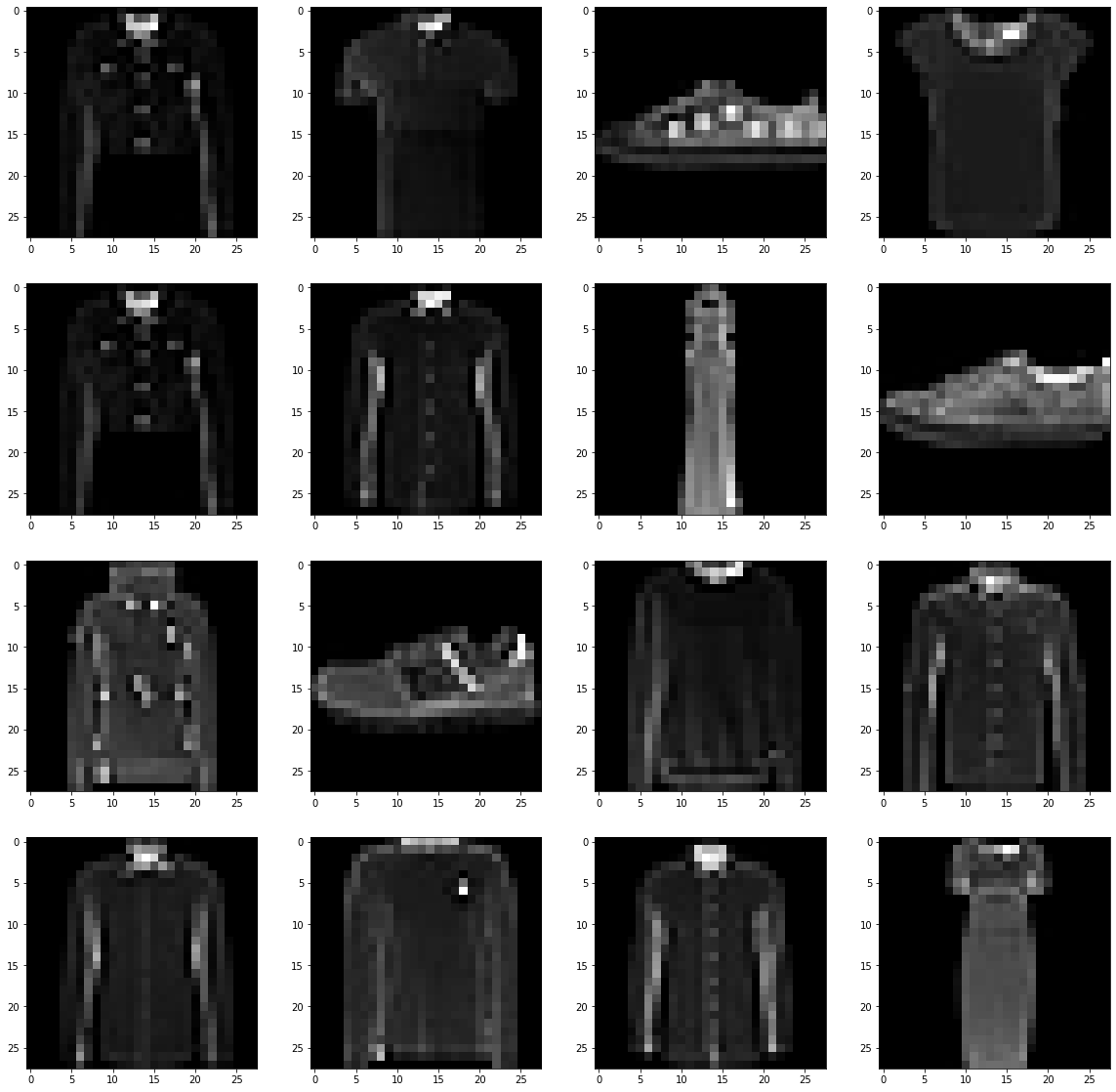}
		\caption{Outliers detected by the binary classifier layer.}
		\label{fig:excludedBCLbig}
	\end{subfigure}
	
	\hfill
	
	\begin{subfigure}[t]{0.3\textwidth}
		\includegraphics[width=\textwidth]{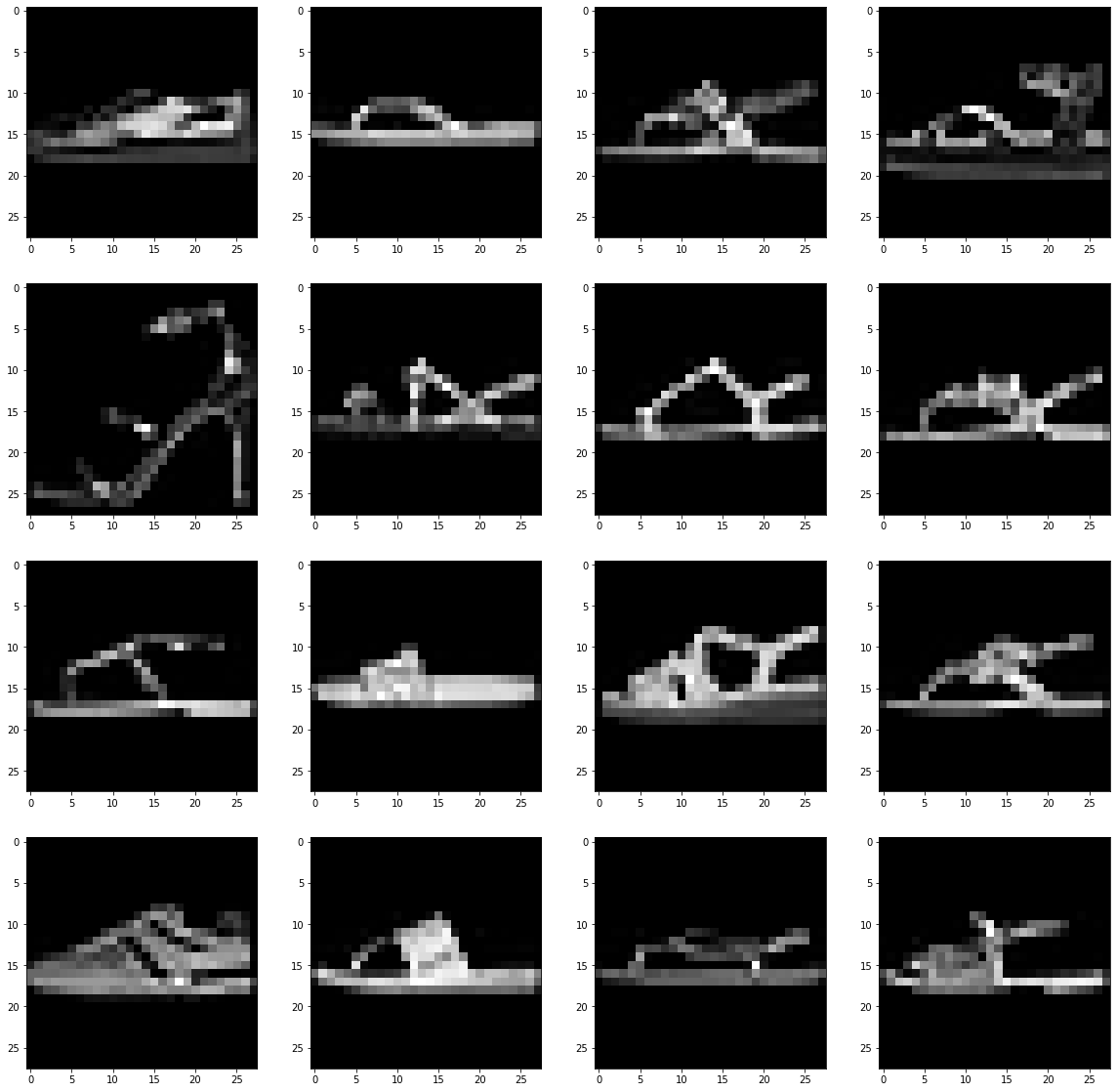}
		\caption{Outliers not detected.}
		\label{fig:notexcudedbig}
	\end{subfigure}
	\caption{Examples of Fashion MNIST outliers detected by the autoencoder (Subfig. \ref{fig:excludedAELbig}), the binary classifier with a certainty threshold of 0.5 (Subfig. \ref{fig:excludedBCLbig}), and not detected (Subfig. \ref{fig:notexcudedbig}).}
	\label{fig:examples_pictures}
\end{figure}

In Figure \ref{fig:examples_pictures}, observe the differences between true positive outliers filtered by the autoencoder and by the binary classifier. Also compare the detected outliers with the false negatives (unfiltered outliers). The outlier images  not detected by the cascade watchdog, such as those seen in Figure \ref{fig:notexcudedbig}, resemble features from the in-distribution MNIST data: Angles, ovals, and writing strokes (see Figure \ref{fig:indistribution}). Between the detected outliers in Figures \ref{fig:excludedAELbig} and \ref{fig:excludedBCLbig}, the differences with the MNIST data are more evident. The outliers detected by the binary classifier (see Figure \ref{fig:excludedBCLbig}) are visually different from the MNIST data, but they also can be perceived as having some subtle features in common. On the other hand, between the outliers detected by the autoencoder (see Figure \ref{fig:excludedAELbig}) it is rare to observe features similar to the in-distribution MNIST dataset.

\section{Conclusion}
The cascade watchdog improves the trade-off between true and false negatives of the stand-alone autoencoder. Adding the binary classifier improves the true positive rate while reducing the false positive rate. The enhancement of the outlier detection is possible due to the production of an augmented dataset by means of adversarial training. The autoencoder, the GAN, and the binary classifier, work in conjunction to produce successful results. 

The results of the experiments show that the binary classifier constitutes a  significant contribution for out-of-distribution identification, encouraging the application of the binary classifier together with the autoencoder in future implementations. The high true positive rate obtained in the experiments, combined with the low false positive rate, confirms that the idea of splitting the out-of-distribution space into different subsets is a good approach for the detection of outliers. While the first layer defense (autoencoder) is capable of detecting most of the outliers, the second layer defense specializes in the out-of-distribution subspace that is closer to the manifold of the distribution. Adversarial training is capable of generating an augmented dataset to train the binary classifier. The cascade watchdog multi-tiered adversarial guard passes the proof of concept stage successfully and has the potential to be applied more broadly to real world classification problems, which require the system to identify out-of-distribution inputs.

\bibliographystyle{bst/sn-basic}
\bibliography{sn-bibliography}


\end{document}